\documentclass[journal,twoside,print]{ieeecolor}

\usepackage{generic}
\usepackage{cite}
\usepackage{amsmath,amssymb,amsfonts}
\usepackage{algorithmic}
\usepackage{graphicx}
\usepackage{textcomp}
\usepackage[acronym]{glossaries}
\usepackage{placeins}
\usepackage{booktabs}
\usepackage{array, multicol, multirow}
\usepackage{adjustbox}
\usepackage{tabularray}
\usepackage{threeparttable}
\usepackage{ifthen}

\makeatletter
\let\NAT@parse\undefined
\makeatother
\usepackage{hyperref}

\newboolean{showcolor}
\setboolean{showcolor}{false}
\newcommand{\bluetext}[1]{%
  \ifthenelse{\boolean{showcolor}}%
    {\textcolor{blue}{#1}}%
    {#1}%
}
\newcommand{\blue}[1]{%
  \ifthenelse{\boolean{showcolor}}%
    {\color{blue}{#1}}%
    {#1}%
}

\newacronym{TMS}{TMS}{transcranial magnetic stimulation}
\newacronym{RI-TMS}{RI-TMS}{robot-assisted image-guided TMS}
\newacronym{RA-TMS}{RA-TMS}{robot-assisted TMS}
\newacronym{Robo-TMS}{Robo-TMS}{robot-assisted TMS}
\newacronym{Robo-mTMS}{Robo-mTMS}{robot-assisted mTMS}
\newacronym{nTMS}{nTMS}{navigated TMS}
\newacronym{rTMS}{rTMS}{repetitive TMS}
\newacronym{dTMS}{dTMS}{deep TMS}
\newacronym{mTMS}{mTMS}{multi-locus TMS}
\newacronym{DECS}{DECS}{direct electrical cortical stimulation}
\newacronym{TES}{TES}{transcranial electric stimulation}
\newacronym{MRI}{MRI}{magnetic resonance imaging}
\newacronym{fMRI}{fMRI}{function MRI}
\newacronym{FEM}{FEM}{finite element method}
\newacronym{BEM}{BEM}{boundary element method}
\newacronym{FDM}{FDM}{finite difference method}
\newacronym{IBEM}{IBEM}{inverse boundary element method}
\newacronym{OCD}{OCD}{obsessive-compulsive disorder}
\newacronym{FDA}{FDA}{Food and Drug Administration}
\newacronym{NIMH}{NIMH}{National Institute of Mental Health}
\newacronym{E-field}{E-field}{electric field}
\newacronym{HCA-coil}{HCA-coil}{halo circular assembly coil}
\newacronym{M1}{M1}{primary motor cortex}
\newacronym{MT}{MT}{motor threshold}
\newacronym{MEP}{MEP}{motor evoked potential}
\newacronym{LED}{LED}{light-emitting diode}
\newacronym{IR}{IR}{infrared}
\newacronym{FOV}{FOV}{field of view}
\newacronym{GPU}{GPU}{graphics processing unit}
\newacronym{CT}{CT}{computed tomography}
\newacronym{VAS}{VAS}{visual analogue scale}

\begin{document}
\bstctlcite{IEEEexample:BSTcontrol}

\title{Robot-assisted Transcranial Magnetic Stimulation (Robo-TMS): A Review}
\author{
Wenzhi Bai$^{1}$, 
Andrew Weightman$^{1}$, 
Rory J.~O'Connor$^{2}$,
Zhengtao Ding$^{1}$,
Mingming Zhang$^{3}$,
Sheng Quan Xie$^{2}$,
and Zhenhong Li$^{1,\dag}$
\vspace{1mm}\\
$^{1}$University of Manchester\quad
$^{2}$University of Leeds\quad
$^{3}$Southern University of Science and Technology
\thanks{IEEE Transactions on Neural Systems and Rehabilitation Engineering.}
\thanks{DOI: \href{https://ieeexplore.ieee.org/document/11071331}{10.1109/TNSRE.2025.3585651}}
\thanks{$^{\dag}$ Corresponding Author (email:zhenhong.li@anchester.ac.uk)}
}
\maketitle

\begin{abstract}
\Gls{TMS} is a non-invasive and safe brain stimulation procedure with growing applications in clinical treatments and neuroscience research. However, achieving precise stimulation over prolonged sessions poses significant challenges. By integrating advanced robotics with conventional \gls{TMS}, \gls{Robo-TMS} has emerged as a promising solution to enhance efficacy and streamline procedures. Despite growing interest, a comprehensive review from an engineering perspective has been notably absent. This paper systematically examines four critical aspects of \gls{Robo-TMS}: hardware and integration; calibration and registration; neuronavigation systems; and control systems. We review state-of-the-art technologies in each area, identify current limitations, and propose future research directions. Our findings suggest that broader clinical adoption of \gls{Robo-TMS} is currently limited by unverified clinical applicability, high operational complexity, and substantial implementation costs. Emerging technologies—including marker-less tracking, non-rigid registration, learning-based \acrfull{E-field} modelling, individualised \acrfull{MRI} generation, robot-assisted \acrlong{mTMS} (\acrshort{Robo-mTMS}), and automated calibration and registration—present promising pathways to address these challenges. 

\end{abstract}

\begin{IEEEkeywords}
Transcranial Magnetic Stimulation, Non-invasive Brain Stimulation, Medical Robots, Image-Guided Robotic Systems, Registration, Neuronavigation, \blue{Optical Tracking Systems}
\end{IEEEkeywords}

\section{Introduction} \label{introduction}
\IEEEPARstart{T}{ranscranial} magnetic stimulation (\acrshort{TMS}) is a non-invasive and safe brain stimulation procedure first introduced in 1985 \cite{barker1985noninvasive}. It utilises a time-varying electromagnetic field generated by coils to induce an electric current in targeted brain regions \cite{groppa2012practical, siebner2022transcranial, edwards2024practical}. As neurogenic diseases and mental disorders remain leading causes of disability and continue to pose a significant global health burden \cite{friedrich2017depression}, there is a growing demand for effective diagnostic and therapeutic tools. Leveraging its neurostimulation and neuromodulation capabilities, \gls{TMS} has been extensively explored for diagnosing and treating various neurogenic diseases and mental disorders \cite{rossi2021safety, dilazzaro2021diagnostic, vucic2023clinical}, as well as for brain mapping research \cite{haddad2020preoperative, giuffre2021reliability, kahl2022active, kahl2023reliability}. Since \gls{TMS} does not require active subject participation, it can be effectively performed on individuals who are paralysed, sedated, or uncooperative \cite{ruohonen2010navigated}. Following its initial approval by the U.S. \gls{FDA} in 2008, \gls{TMS} has seen growing adoption in neurosurgery planning and treating depression, \gls{OCD}, migraines, and smoking cessation especially when standard treatments have proven ineffective \cite{horvath2011transcranial, cohen2022visual}.

\begin{figure}[!htb]
\centering
\includegraphics[width=0.48\textwidth]{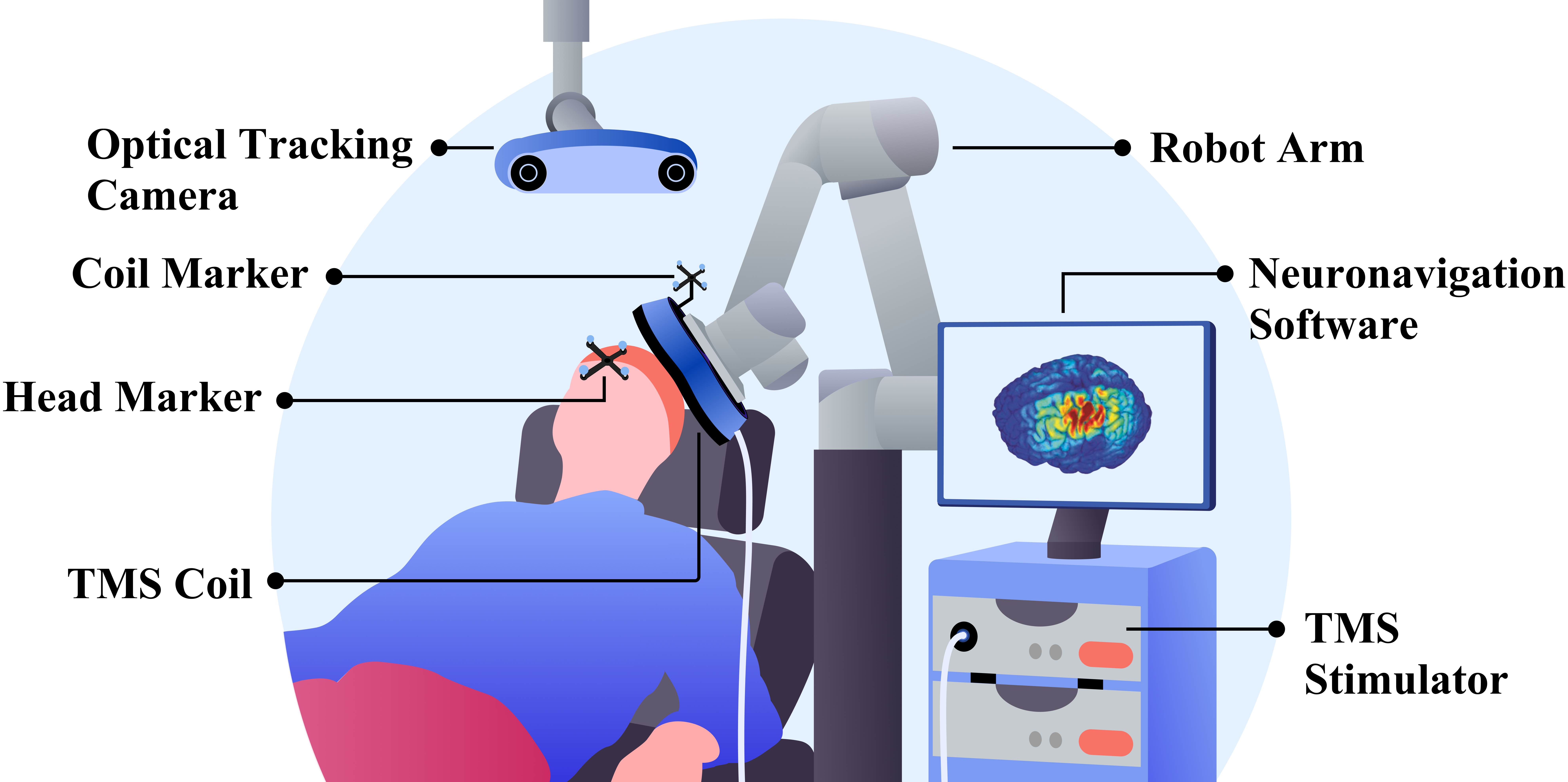}
\centering
\caption{A typical application scenario of \gls{Robo-TMS}. \protect\blue{The subject lies in a chair with a head marker attached. An optical tracking camera monitors head and coil markers to guide the robot arm in compensating for head movements, maintaining a consistent coil-to-head transformation for accurate stimulation. The \gls{TMS} stimulator generates current pulses into the \gls{TMS} coil, and the resulting induced \gls{E-field} within the subject's head is visualised by the neuronavigation software on the monitor.}}
\label{fig:Robo-TMS Scenario}
\end{figure}

In conventional \gls{TMS}, clinicians cannot directly visualise the stimulation spot beneath the scalp, and even the corresponding location on the subject’s head is often obscured by the stimulation coil. Maintaining precise targeting throughout prolonged sessions is further complicated by the need to manually hold the coil. To address these challenges, advanced robotics innovations have been integrated into conventional \gls{TMS} to enhance clinical outcomes. These advancements include neuronavigation systems for accurate stimulation targeting, \gls{E-field} modelling for the estimation of induced \gls{E-field} distribution, optical tracking systems for subject's head movements compensation, and robot arms for stable and precise coil placement. Reflecting these new capabilities, the terminology used to describe these procedures has evolved: \gls{TMS} enhanced with neuronavigation systems is often referred to as \gls{nTMS} \cite{tarapore2016safety}; automated systems that incorporate robot arms are termed robotic \gls{TMS}, robotised \gls{TMS}, \gls{RA-TMS}, or \gls{RI-TMS} \cite{kantelhardt2010robotassisted}. To ensure clarity and consistency, we adopt the term \acrfull{Robo-TMS} to describe the integrated system illustrated in Fig. \ref{fig:Robo-TMS Scenario}.

Recent literature on PubMed indicates a growing interest among medical researchers in more advanced \gls{TMS} devices to meet evolving clinical needs. However, comprehensive reviews on \gls{Robo-TMS}, particularly from an engineering perspective, are absent. Existing reviews primarily focus on the fundamental aspects of \gls{TMS} \cite{wagner2007noninvasive, ruohonen2010navigated, siebner2022transcranial} and its clinical applications \cite{haddad2020preoperative, dilazzaro2021diagnostic, vucic2023clinical}. Some reviews emphasise the importance of neuronavigation systems and the necessity for high accuracy \cite{cleary2010imageguided, smith201630, pivazyan2022basis}, while others delve into specific aspects like \gls{E-field} modelling \cite{perez-benitez2023review, park2024review}, brain mapping \cite{gao2024individualized}, and accuracy comparisons across different systems \cite{herwig2001transcranial, sparing2008transcranial, fang2019current, jeltema2021comparing}. \bluetext{Although early insights into \gls{Robo-TMS} are provided by a book \cite{richter2013robotized} and a doctoral thesis \cite{matthaus2008robotic} over a decade ago, the rapid advancements in robotics underscore the need for a comprehensive review that introduces emerging engineering tools for addressing current challenges in \gls{TMS} practice and informs the development of technology that aligns with clinical requirements.}


This review focuses on the core technologies and components of \gls{Robo-TMS}, while omitting elements that have been widely discussed in conventional \gls{TMS}, such as pulse waveform, cooling, power supply and etc. The main objectives are
\begin{enumerate}
    \item to identify the status and challenges in four critical aspects: hardware and integration, calibration and registration, neuronavigation systems, and control systems;
    \item to bridge the gap between the medical and engineering perspectives, providing a comprehensive understanding of \gls{Robo-TMS};
    \blue{
    \item to propose directions for further research and development by discussing how to potentially enhance clinical outcomes and reduce operational costs.}
\end{enumerate}

To the best of our knowledge, this is the first review to comprehensively survey the status and challenges of \gls{Robo-TMS} from an engineering perspective. We hope it can foster collaboration among clinicians, engineers and researchers, ultimately driving future advancements in the field.

\renewcommand\arraystretch{1.2}
\begin{table*}[!htb]
    \bluetext{
    \caption{The Comparison of Typical \gls{TMS} Systems}
    }
    \centering
    \begin{threeparttable}
    \resizebox{\linewidth}{!}{ 
        \begin{tabular}{cccc}
        \toprule[2pt]
            & \textbf{Conventional \gls{TMS} System} & \textbf{Industrial Robot-based \gls{Robo-TMS} System} & \textbf{Specialised Robot-based \gls{Robo-TMS} System} \\ [1ex]
            \hline
            \textbf{Holders} & Passive Mechanical Holder & Active Industrial Robot Arm & Active Specialised Robot Arm \\
            \textbf{Force / Torque Sensors} & None & Required & Required \\
            \textbf{Optical Tracking} & Optional & Required & Required \\
            \textbf{E-field Modelling} & Optional & Recommended & Recommended \\
            \blue{\textbf{Accuracy\tnote{*}}} & \blue{\(\sim6 \mathrm{mm} / \sim3^\circ\)\cite{ginhoux2013custom, noccaro2021development, xygonakis2024transcranial}} & \blue{\(\sim2 \mathrm{mm} / \sim1.5^\circ\)\cite{ginhoux2013custom, noccaro2021development, xygonakis2024transcranial}} & \blue{\(\sim2 \mathrm{mm} / \sim1.5^\circ\)\cite{ginhoux2013custom, noccaro2021development, xygonakis2024transcranial}} \\
            \blue{\textbf{Contact Force\tnote{*}}} & \blue{\(6-10 \mathrm{N}\) \cite{zakaria2012forcecontrolled}} & \blue{\(\sim2.5 \mathrm{N}\) \cite{lebosse2006robotic, noccaro2021development}} & \blue{\(\sim2.5 \mathrm{N}\) \cite{lebosse2006robotic, noccaro2021development}} \\
            \textbf{Examples} & \adjustbox{valign=m}{\includegraphics[width=0.22\textwidth]{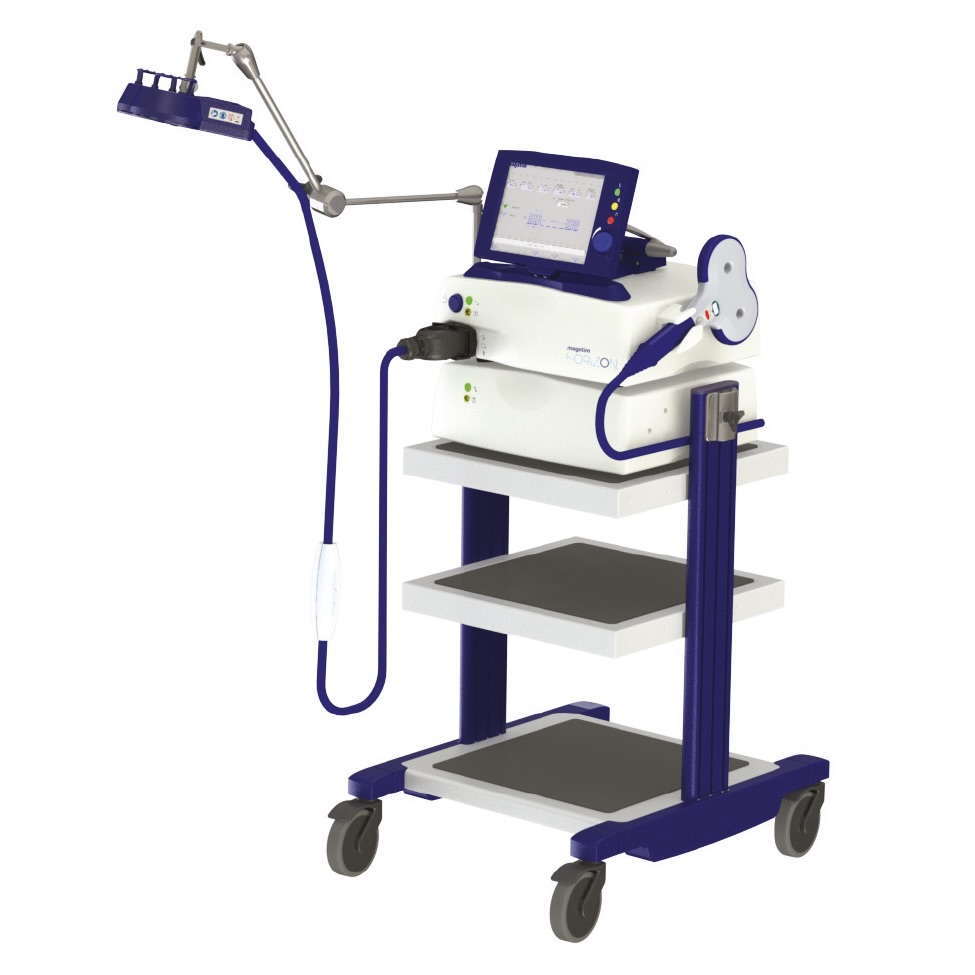}} & \adjustbox{valign=m}{\includegraphics[width=0.20\textwidth]{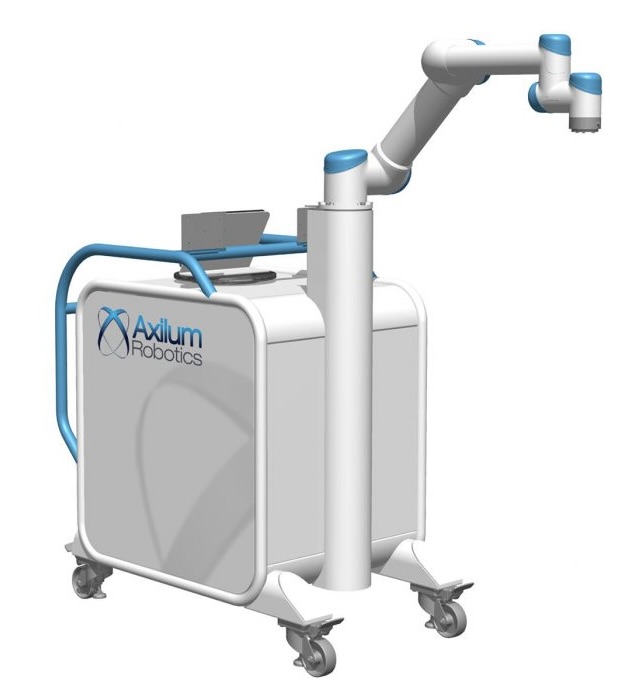}} & \adjustbox{valign=m}{\includegraphics[width=0.18\textwidth]{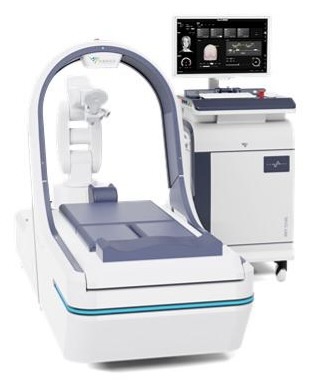}} \\
            & Magstim\textsuperscript{TM} Horizon Lite \cite{magstim} & Axilum Robotics\textsuperscript{TM} TMS-Cobot \cite{axilum} & Yiruide\textsuperscript{TM} Mag-aim \cite{yiruide} \\ 
        \bottomrule[2pt]
        \end{tabular}
    }
    \begin{tablenotes}
        \blue{
        \item[*] Accuracy and contact force listed in the table are for specific commercial systems and experimental prototypes.}
    \end{tablenotes}
    \end{threeparttable}
    \label{tab:comparison of tms conf}
\end{table*}

\section{Hardware and Integration}
\ifthenelse{\boolean{showcolor}}
{
\definecolor{subsectioncolor}{rgb}{0,0,1}
\subsection{System Setup}
\definecolor{subsectioncolor}{rgb}{0,0.541,0.855}
}
{\subsection{System Setup}}
Manually holding the coil for prolonged \gls{TMS} procedures not only exhausts clinicians but also leads to inaccuracies caused by fatigue, affecting stimulation efficacy. While mechanical holders offer some stability \cite{chronicle2005development}, they are unable to compensate for unexpected head movements. Compromised solutions, such as instructing subjects to remain still or employing head restraints, often cause discomfort and stress. \gls{Robo-TMS}, which integrates neuronavigation with a robot arm to ensure accurate coil targeting and precise coil placement, has been developed aiming to enhance both comfort and the overall efficacy of stimulation. In 2000, the first \gls{Robo-TMS} system was developed by mounting a \gls{TMS} coil onto the neurosurgical robot NeuroMate, a five-joint serial robot \cite{narayana2000use}. This system achieves precise coil placement with an accuracy of approximately \(2 \mathrm{mm}\) \cite{lancaster2004evaluation}. However, due to the lack of head-tracking capability, it requires subjects to maintain a stationary head position during procedures.

Subsequent \gls{Robo-TMS} systems can be broadly categorised into two types: industrial robot-based systems, valued for their adaptability and accessibility; and specialised robot-based systems, tailored to clinical safety and efficacy requirements with enhanced integration. A typical industrial robot-based system is first introduced by adapting a standard six-joint industrial robot and integrating it with an optical tracking system for automated neuronavigation \cite{matthaus2006robotized, matthaus2008robotic}. Later advancements incorporate force sensors for collision avoidance and safe coil-to-head contact \cite{richter2013robotized}. Other industrial robot-based systems employ similar configurations \cite{g.pennimpede2013hot, todd2014brain}, including a parallel industrial robot-based system designed to enhance stiffness and reduce moving mass, enabling faster and more precise coil placement \cite{j.j.dejong2012method}. In parallel, specialised robot-based systems also incorporate the optical tracking system and force sensors but differ by employing customised designs to stabilise the dynamic coil cable and constrain coil movement tangentially to the subject's head, thereby better meeting clinical requirements for safety and efficacy \cite{lebosse2006robotic, lebosse2007robotic, zorn2012design, ginhoux2013custom}. Both types of systems underscored the importance of integrating force or torque sensors to adjust contact forces dynamically \cite{xiangyi2010design}, ensuring the stimulation coil maintained optimal proximity to the scalp—a key factor in achieving effective stimulation \cite{zangen2005transcranial}. \bluetext{A comparison of typical \gls{TMS} system configurations and their performance is provided in TABLE \ref{tab:comparison of tms conf}.}

Building on these typical systems, innovations have continued to refine \gls{Robo-TMS}. To address tracking loss caused by robot occlusion, some systems mount cameras directly onto the robot's end-effector, ensuring continuous visual feedback during procedures \cite{liu2022out}. Teleoperated haptic-enabled \gls{Robo-TMS} systems are developed to facilitate remote \gls{TMS} treatments, providing significant benefits for rural healthcare delivery \cite{kebria2023hapticallyenabled}. Recent advancements have also focused on marker-less tracking, eliminating the need for head and coil markers \cite{z.xiao2018cortexbot, chen2020noattachment}. This method simplifies registration processes, minimises manual errors, shortens treatment times, and consequently enhances the overall experience for both clinicians and subjects. Additionally, novel \gls{mTMS} systems utilise coil arrays to produce spatially controlled magnetic fields \cite{koponen2018multilocus, nieminen2022multilocus}. By integrating robotic movements with spatially controlled magnetic fields, \gls{Robo-mTMS} achieves rapid and accurate targeting \cite{matsuda2024robotic, sinisalo2024modulating}.

Currently, four commercial \gls{Robo-TMS} systems lead the market: Axilum Robotics' TMS-Cobot and ANT's Smartmove (industrial robot-based systems); Axilum Robotics' TMS-Robot and Yiruide's Mag-aim (specialised robot-based systems). Notably, Axilum Robotics' TMS-Cobot is the first \gls{Robo-TMS} system to receive clearance from the U.S. \gls{FDA} for clinical use. Although \gls{Robo-TMS} offers advantages over conventional \gls{TMS}, its clinical adoption remains in the early stages and continued developments are essential to fully realise its potential.

\subsection{Coil Design}\label{coil design}
\begin{figure}[!htb]
\centering
\includegraphics[width=0.4\textwidth]{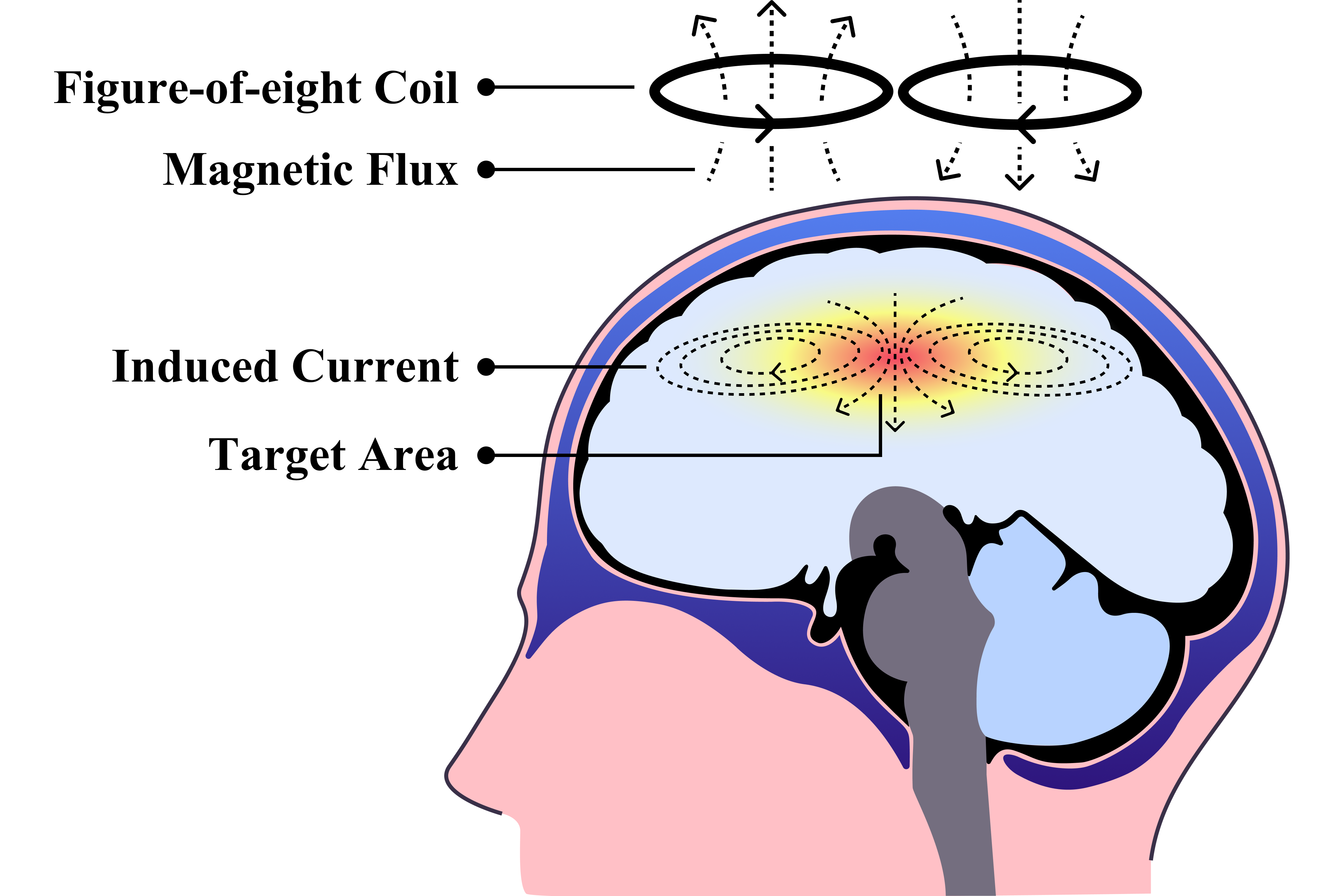}
\caption{The principle of \gls{TMS}. \gls{TMS} operates on the principle of electromagnetic induction (Faraday's law of induction). A rapidly changing current passing through the coil generates a brief magnetic field pulse, which induces electric currents in the targeted brain region. The distribution of the induced \gls{E-field} is influenced by the coil design, its placement on the scalp, and the intensity of the applied current.}
\label{fig:electromagnetic induction}
\end{figure}

The \gls{TMS} uses a high-voltage power supply to charge capacitors, which are then rapidly discharged into the \gls{TMS} coil. This process generates a brief magnetic field pulse, delivered via electromagnetic coils placed on the subject’s scalp. The pulse induces electric currents in the underlying cortical tissue, which can stimulate neurons when these currents exceed a threshold \cite{zhang2021theoretical}. This capability allows \gls{TMS} to stimulate specific brain regions, potentially controlling neural activation. The principle of \gls{TMS} is shown in Fig. \ref{fig:electromagnetic induction}.

The design of the \gls{TMS} coil, including its shape, size, and number of winding turns, is crucial for stimulation safety and efficacy. The coil’s configuration \bluetext{affects} not only how the \gls{E-field} is distributed within the brain but also its intensity and pulse width. The primary objective of coil design is to enable targeted brain stimulation while minimizing unintended activation of surrounding brain regions \cite{rossi2021safety, siebner2022transcranial}.

\subsubsection{Standard Coils}
The circular coil or round coil is the most basic coil design, generating an annular \gls{E-field}. Circular coils, typically \(9 \mathrm{cm}\) in diameter, penetrate relatively deep, inducing currents over broad brain regions. This non-focal stimulation uniformly activates all regions beneath the coil’s annulus at peak current densities. As a result, effective stimulation requires placing the annulus, rather than the coil's centre, over the target cortex \cite{groppa2012practical, deng2013electric}.

The figure-of-eight coil, first introduced in 1988 \cite{ueno1988localized} and also known as a double or butterfly coil, consists of two adjacent circular coils with opposing currents. This design generates the highest current density directly beneath the coil intersection, where the \gls{E-field} aligns parallel to the wires at the centre. Consequently, focal stimulation is achieved by placing the coil's centre over the target, concentrating the current in a central region with a density two to three times higher than at the edges. This focality makes the figure-of-eight coil ideal for precise cortical stimulation in both research and clinical applications \cite{groppa2012practical, deng2013electric, peterchev2015advances}.

\subsubsection{Coils for Deep TMS}
Standard \gls{TMS} coils, mainly stimulate superficial cortical regions, as their \gls{E-field} intensity decreases rapidly with depth \cite{bagherzadeh2019effect}. Consequently, reaching deeper brain regions necessitates substantially higher intensities, which may exceed the capabilities of standard coils and pose safety concerns such as discomfort and side effects. To address these limitations, \gls{dTMS} coils have been developed with optimised \gls{E-field} distribution that has higher intensity at deeper regions while minimising excess stimulation at the cortical surface by reducing \gls{E-field} decay \cite{lu2017comparison}. This innovation is particularly valuable for treatments requiring direct stimulation of deep neural pathways. Below are some common \gls{dTMS} coil designs:

\paragraph{Double-cone Coils}
The double-cone coil, also known as an angled butterfly coil, is an enlarged version of the figure-of-eight coil. Its two circular windings are angled towards the subject’s head, enhancing the magnetic field strength and electrical efficiency at greater depths. This coil penetrates deeper into the brain but is less focal than the standard figure-of-eight coil, making it suitable for stimulating regions \(3-4 \mathrm{cm}\) deep, such as the primary motor region of the leg \cite{deng2013electric}.

\paragraph{H-coils}
The H-coil, or termed Hesed coil, is designed to induce a deeper \gls{E-field} than figure-of-eight coils, albeit with reduced focality \cite{zangen2005transcranial}. The H-coil features a complex winding pattern and larger dimensions, with elements strategically placed around the target brain region to generate a summation of the \gls{E-field} at a depth of \(4-6 \mathrm{cm}\). This design allows for deeper brain stimulation without excessively stimulating the cortical surface \cite{fadini2009hcoil, deng2013electric, lu2017comparison}.

\paragraph{Halo Circular Assembly Coils}
The \gls{HCA-coil} is a large circular coil designed to be placed around the head, providing sub-threshold \gls{E-field} stimulation in deep brain tissues. It can be used in conjunction with a conventional circular coil positioned at the top of the head. This design allows for more flexible and deeper brain stimulation than standard circular coils, with additional coaxial circular coils developed to reduce \gls{E-field} intensity in superficial cortical regions \cite{deng2013electric}.

\paragraph{Helmet Coils}
The helmet coil, a \gls{dTMS} coil with customised geometry, is designed using continuous current density \gls{IBEM} to optimise depth control and minimise power dissipation. Its subject-specific design, tailored to the shape of the subject’s head, enables focal stimulation in regions such as the prefrontal cortex and right temporal lobe. According to \cite{membrilla2024design}, helmet coils can increase stimulation depth by over 15\% compared to H-coils with similar focality.

\subsubsection{Coils for Multi-locus TMS}
Conventional \gls{TMS} coils stimulate a fixed location directly beneath the coil, requiring physical movement to target different brain regions. This physical movement is slow and limits applications like feedback-controlled stimulation, where rapid adjustments are necessary. To overcome these limitations, a 2-coil \gls{mTMS} system has been developed, enabling electronic targeting of nearby cortical regions along a \(30 \mathrm{mm}\) line segment without moving the coil \cite{koponen2018multilocus}. Further advancements include a 5-coil \gls{mTMS} system that allows for adjusting the location and orientation of the \gls{E-field} maximum within a \(30 \mathrm{mm}\) diameter cortical region \cite{nieminen2022multilocus}. Design optimisation of \gls{mTMS} has also been investigated to address the manufacturing challenges of \gls{mTMS} coils, such as excessive winding density \cite{rissanen2023advanced}. Although \gls{mTMS} coils enable faster and more complex stimulation patterns, targeting multiple brain regions with varying \gls{E-field} directions, timings, and intensities, the stimulation region of \gls{mTMS} is still limited due to manufacturing challenges.

\begin{figure}[!htb]
\centering
\includegraphics[width=0.4\textwidth]{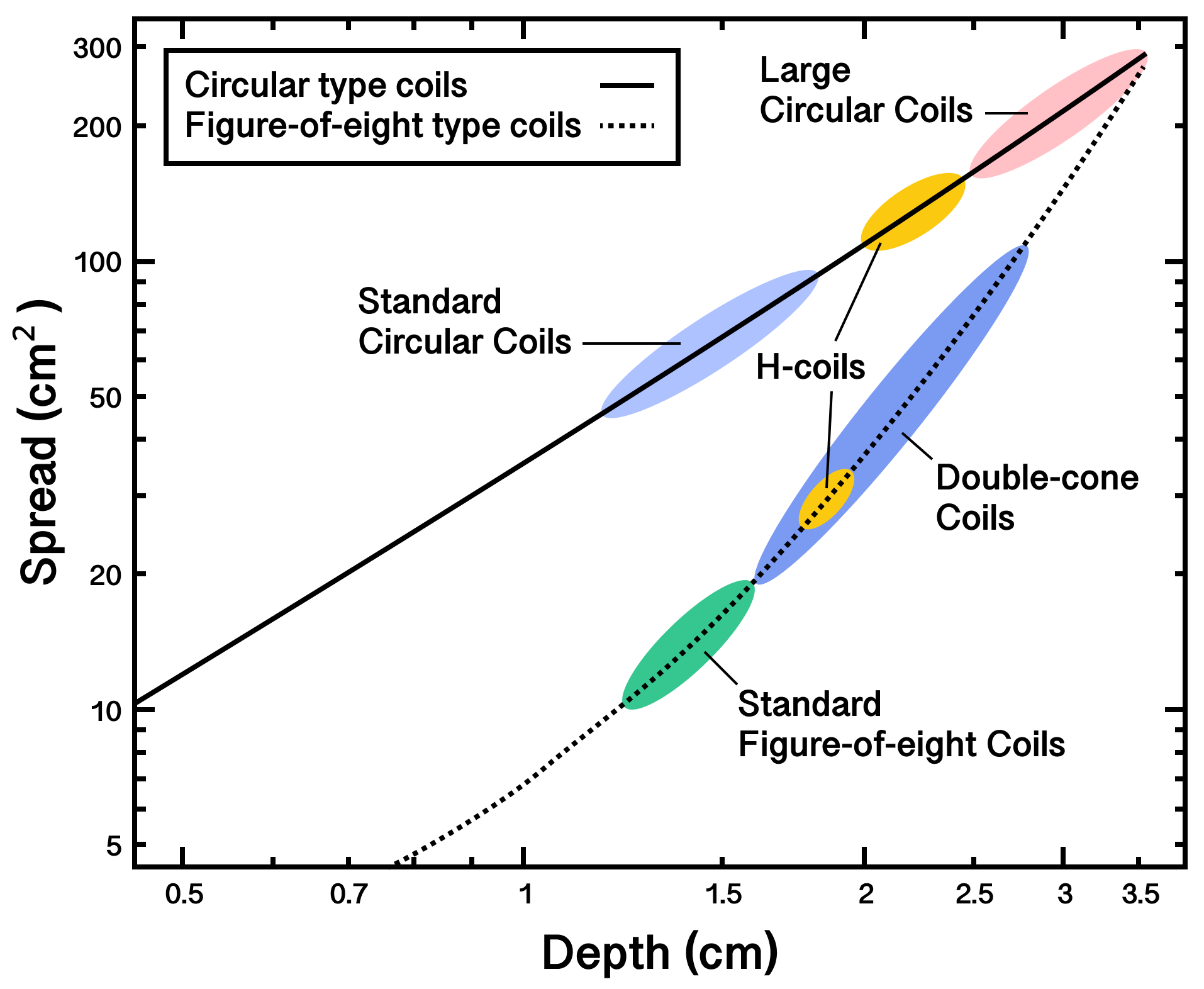}
\caption{The depth-focality trade-off in \gls{TMS} coil design. The depth is quantified by \(d_{1/2}\), and the focality is quantified by the spread \(S_{1/2}\). For any type of coils, deeper stimulation is accompanied by increased spread, indicating reduced focality \cite{deng2013electric, peterchev2015advances}.}
\label{fig:coil design trade-off}
\end{figure}

\subsubsection{Depth-focality Trade-off}
In \gls{TMS} coil design, there is a fundamental trade-off between stimulation depth and focality as shown in Fig. \ref{fig:coil design trade-off}. Larger coils penetrate deeper into the brain but at the cost of reduced focality, stimulating larger regions of the brain surface. Conversely, coils designed for focal surface stimulation experience rapid \gls{E-field} decay at greater depths, limiting their efficacy in deeper regions \cite{bagherzadeh2019effect}. Simulations of 50 coil designs highlight this depth-focality trade-off, illustrating the challenge of achieving both deep penetration and precise targeting in \gls{TMS} \cite{deng2013electric}.

This trade-off can be quantified by metrics such as \(E_{max}\) (the maximum \gls{E-field} strength), \(d_{1/2}\) (the depth at which the \gls{E-field} drops to half of \(E_{max}\)), and \(S_{1/2}=\frac{V_{1/2}}{d_{1/2}}\) (a focality measure), where \(V_{1/2}\) indicates the volume beneath the cortical surface in which the \gls{E-field} exceeds \(\frac{1}{2}E_{max}\), with smaller \(S_{1/2}\) values indicating better focality \cite{thielscher2004electric, deng2013electric, drakaki2022database}. Based on quantitative metrics, multi-objective optimisation algorithms can be used to generate the computational design of \gls{TMS} coils that balance focality, depth, and energy efficiency, reaching specific depths without exceeding predefined \gls{E-field} strength limits \cite{gomez2018design}. A similar trade-off also applies to \gls{mTMS}. By examining the interplay among coil number, focality, and the cortical region over which the \gls{E-field} peak can be controlled, researchers have shown that expanding the controllable region while maintaining the same number of coils necessitates a compromise in E-field focality \cite{nurmi2021tradeoff}.

\section{Calibration and Registration}\label{calibaration and registration}
Calibration and registration are essential to ensure accurate targeting and efficient stimulation of brain regions by establishing the initial settings and subject-specific configurations, including workspace calibration, registration and stimulation intensity calibration. These processes align the robot’s workspace, optical tracking systems, and the subject’s anatomy, while also adapting the equipment’s parameters to individual characteristics. 

\subsection{Workspace Calibration}
Workspace calibration in \gls{Robo-TMS} synchronises the coordinate systems of the camera and the robot, enabling accurate transformation of visual information into robot movements. Without proper calibration, the robot cannot accurately interpret the target within its workspace, leading to operational inaccuracies.

In robotics, calibrating the spatial relations between a robot arm and a camera is a well-established problem, commonly known as hand-eye or robot-world calibration. The calibration can be divided into two categories  based on the configurations of the camera-robot system: eye-to-hand and eye-in-hand. \bluetext{As a natural extension of conventional \gls{TMS} systems where the coil is handheld, most \gls{Robo-TMS} systems \cite{richter2011robust, wang2018nonorthogonal, chen2020noattachment, noccaro2021development} adopt the same eye-to-hand configuration to keep the coil-holding device lightweight and mechanically simple. In this setup, the camera is fixed in the workspace and operates independently of the robot arm’s movements. More recent \gls{Robo-TMS} systems \cite{liu2022out} explore the eye-in-hand configuration, where the camera is mounted on the end-effector and moves with the robot arm. This setup ensures that the camera’s line of sight remains unobstructed by the robot throughout the procedure. However, it is generally limited to marker-based tracking systems, as mounting the camera on the end-effector makes it difficult to capture facial features, which are commonly used in marker-less tracking.}

Regardless of the configuration, the core concept of hand-eye calibration is to ensure that the relative motion observed by both the robot (hand) and the camera (eye) is consistent, allowing the rigid transformation between different components to be accurately estimated through multiple observations. Mathematically, the common eye-to-hand calibration problem can be expressed by the equation shown below.

\begin{equation}
\mathbf{M}\mathbf{X} = \mathbf{Y}\mathbf{N}
\end{equation}
where all matrices are \(4\times4\) homogeneous transformations in special Euclidean group \(SE(3)\). Here,  \(\mathbf{M}\) represents the transformation between the robot’s end-effector frame and the robot’s base frame,  \(\mathbf{N}\) represents the transformation between the coil frame and the camera’s base frame, and \(\mathbf{X}\) and \(\mathbf{Y}\) are the unknown transformations between the robot's end-effector and coil frames, and the robot's base and camera's base frames, respectively.

Building on standard calibration techniques, the method described in \cite{noccaro2021development} is optimised for \gls{Robo-TMS} by accounting for the robot's specific workspace during treatments, characterised as a spherical shell around the subject’s head. Tested across three calibration algorithms—SGO \cite{ha2016stochastic}, QR24 \cite{ernst2012nonorthogonal}, and QUAT \cite{dornaika1998simultaneous}—it reduces positional error by 34\% and orientation error by 19\%. \bluetext{Additionally, a method employs an optical tracking system to independently measure coil geometry and compute the required transformation with high accuracy \cite{liu2024gbec}.}

\subsection{Registration}
After workspace calibration, the robot arm can accurately interpret and locate targets from the optical tracking system within its workspace. However, to stimulate the target tissue effectively, individual factors—such as the size and shape of the head and brain, the distance between the stimulating coil and the target tissue, as well as the location and orientation of anatomical structures—must be defined for each subject \cite{ruohonen2010navigated, liu2025imageguided}. Consequently, an individual or individualised \gls{MRI} is typically required to represent the subject’s static neuroanatomy for alignment. In fact, spatial and temporal alignment are critical in any multi-source system. In \gls{Robo-TMS}, registration refers to aligning the \gls{MRI} of the subject's neuroanatomy with the physical head in the workspace, a prerequisite for neuronavigation \cite{orringer2012neuronavigation}. Once registration is complete, the subject's neuroanatomy can also be represented in the robot's workspace, enabling \gls{Robo-TMS} to visualise brain structures despite their location within the physical head. This alignment facilitates accurate stimulation within the most intricate organ in the human body.

A comprehensive analysis reveals that the registration method is a major contributor to the level of error in neuronavigation \cite{nieminen2022accuracy}. In \gls{Robo-TMS}, two main types of registration methods are commonly used: landmark-based registration and surface-based registration. Landmark-based registration involves manually selecting at least three corresponding points on the \gls{MRI} and the subject’s head, but it generally offers lower accuracy and precision than surface-based methods due to limited landmarks \cite{matilainen2024verification}. Common landmarks include the nasion (bridge of the nose), the left and right pre-auricular points, and occasionally the tip of the nose. However, the manual selection of landmarks leads to registration errors. In \cite{nieminen2022accuracy}, mean errors of \(2.9 \mathrm{mm} / 1.1^{\circ}\) for accuracy and \(1.4 \mathrm{mm} / 0.6^{\circ}\) for precision were reported, with larger errors observed in posterior regions of the head \cite{shamir2009localization, omara2014anatomical}.

By contrast, surface-based registration, which starts with landmark-based alignment followed by digitising points on the scalp and matching them to a scalp mesh derived from the \gls{MRI}, significantly reduces human-induced errors. This method achieves mean errors of  \(1.0 \mathrm{mm} / 0.7^{\circ}\) for accuracy and \(0.6 \mathrm{mm} / 0.4^{\circ}\) for precision \cite{nieminen2022accuracy}. However, challenges remain, particularly with the use of 3D stylus digitisers, which require manual manipulation and longer measurement times. As highlighted by \cite{smith201630}, the time required for registration is a critical factor in operating room efficiency, as clinicians must wait for registration to be completed before continuing the procedure. During the procedure, any fiducial marker shift necessitates re-registration, further delaying the process. To address these limitations, advanced surface-based 3D digitisation techniques, such as 3D laser scanning \cite{hironaga2019proposal} and photogrammetry-based systems \cite{liu2023transcranial}, have been proposed. These methods enhance accuracy while streamlining the registration process.

\bluetext{
In addition, registration can also be performed using either rigid or non-rigid methods. Rigid registration considers only translation and rotation, ignoring the complex spatial deformations inherent to soft tissues. While rigid methods currently dominate \gls{Robo-TMS} due to the skull’s rigid structure, they are less suitable for aligning soft tissue regions such as the face and brain, where non-rigid deformation is common. Non-rigid registration methods offer the potential to improve alignment accuracy in these areas by modelling such deformations. However, they remain relatively immature and face challenges including sensitivity to noise, outliers, varying deformation levels, and incomplete data \cite{monji2023review}. These limitations reduce robustness and increase computational complexity. Nevertheless, as non-rigid methods mature, they are expected to further improve registration accuracy and precision in \gls{Robo-TMS} applications \cite{cleary2010imageguided, risholm2011multimodal, orringer2012neuronavigation}.}

\subsection{Stimulation Intensity Calibration}
\bluetext{
The \gls{TMS} treatment begins with a mandatory step to determine the appropriate stimulation intensity for the target region \cite{rossi2021safety, vucic2023clinical}, as this depends on the subject’s unique neuroanatomy and is essential for personalised treatment. Typically, the \gls{MT}, defined as the minimum stimulation intensity that corresponds to a 50\% probability of eliciting a motor response, serves as a reference for calibrating stimulation intensity at other cortical targets. To establish the \gls{MT}, clinicians first search within the \gls{M1} using a clinically guided starting intensity to locate the motor hotspot, where stimulation could evoke responses in a target muscle, commonly the first dorsal interosseous or abductor pollicis brevis of the dominant hand. Once the motor hotspot is identified, the stimulation intensity is gradually adjusted to determine the resting or active \gls{MT} by recording the muscle responses evoked by stimulation. Subsequent stimulation intensities for the target region are then expressed as a percentage of the individual’s \gls{MT}, in accordance with safety guidelines and to standardise stimulation \cite{herbsman2009motor, edwards2024practical}.}

While early methods efficiently automate \gls{MT} estimation \cite{awiszus2003tms}, clinicians still need to first manually identify the hotspot, introducing variability between clinicians. To address this, the first automated hotspot localisation method based on \glspl{MT} mapping has been introduced \cite{meincke2016automated}. Building on this, AutoHS—a probabilistic Bayesian model—has been developed to automate hotspot localisation, offering improved reproducibility, speed, and reliability compared to earlier methods \cite{harquel2017automatized}. By integrating AutoHS with \gls{Robo-TMS} and automated \gls{MT} estimation, the first fully automated setup procedure for \gls{Robo-TMS} has been proposed, aiming to reduce inter-subject variability and minimise setup time \cite{harquel2017automatized}.

However, these automated stimulation intensity calibration methods require a large number of \gls{MEP} measurements and operate with a fixed stimulation orientation. Furthermore, these methods heavily rely on sparse grids between coil positions, potentially limiting accuracy. To overcome these limitations, recent studies \cite{tervo2020automated, tervo2022closedloop} introduce algorithms supported by \gls{mTMS}, designed to automatically optimise stimulation parameters, location, and orientation to elicit the largest \glspl{MEP}.

\begin{figure*}[!htb]
\centering
\includegraphics[width=0.9\textwidth]{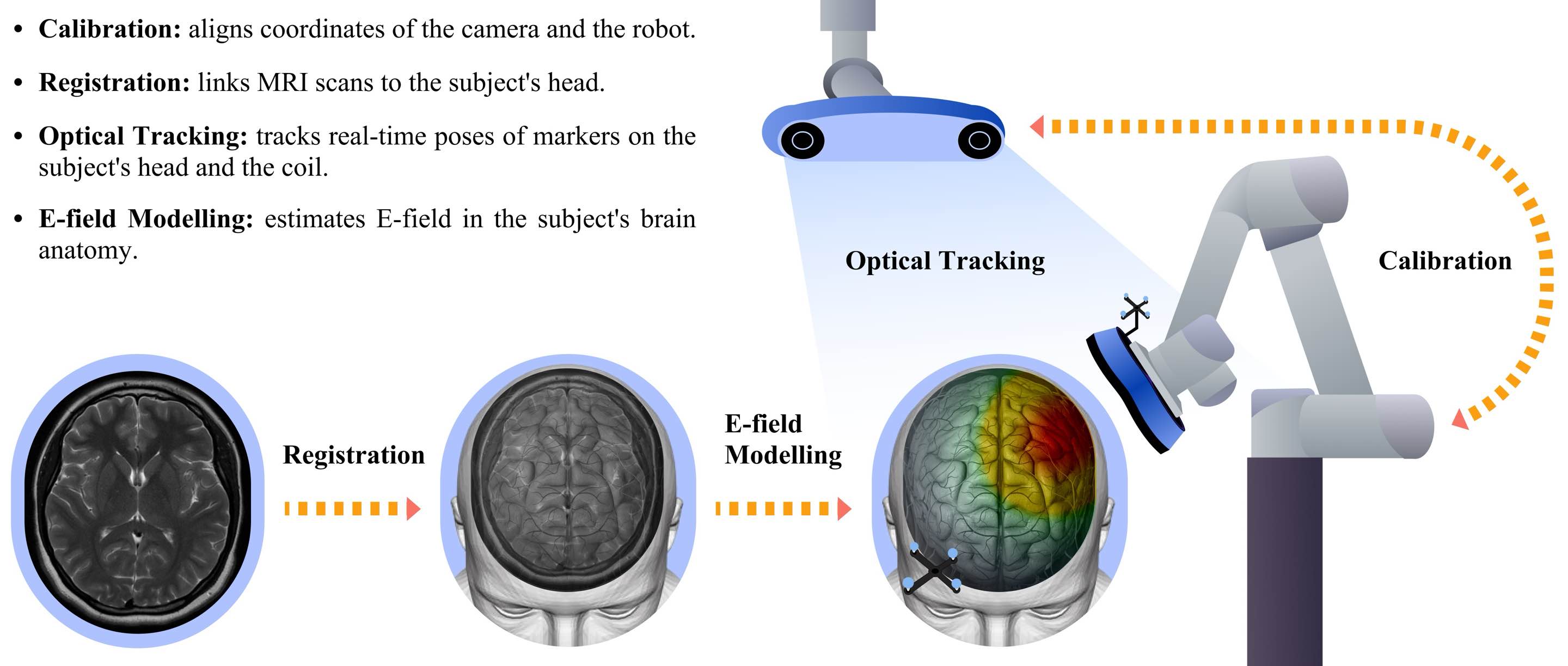}
\caption{The neuronavigation system for \gls{Robo-TMS}. The optical tracking camera continuously monitors the real-time poses of markers on the coil and the subject's forehead, enabling accurate tracking of both the coil and head poses. Through calibration, these poses are transformed from the camera’s base frame to the robot’s base frame for robot control, providing an accurate coil pose relative to the subject's head. With individual \glspl{MRI} registered to the subject’s physical head, \gls{E-field} modelling can then visualise the resulting \gls{E-field} distribution within the brain.}
\label{fig:neuronavigation}
\end{figure*}

\section{Neuronavigation Systems}
\bluetext{The neuronavigation system utilises optical tracking and \gls{E-field} modelling, alongside neuroanatomical data, to provide real-time guidance to clinicians during procedures.} This system serves as a critical component of \gls{Robo-TMS}. \bluetext{As discussed in Section \ref{calibaration and registration}, workspace calibration establishes the transformation between the coordinate systems of the optical tracking camera, the coil, and the robot arm,} while registration establishes the relationship between the image of an individual’s brain anatomy and their physical head. However, compared to calibration and registration that occur during the initial setup, the neuronavigation system is essential for determining the real-time pose relationship between the head and the coil within the optical tracking camera’s coordinate system, as well as the induced \gls{E-field} based on this relative pose during the procedure. An overview of the neuronavigation system for \gls{Robo-TMS} is shown in Fig. \ref{fig:neuronavigation}. The U.S. \gls{NIMH} recommends neuronavigation on an individual basis in all \gls{TMS} applications to standardise coil localisation and account for individual differences in the delivered dose, thereby improving the rigour and repeatability of non-invasive brain stimulation studies \cite{rossi2021safety}.

\subsection{Optical Tracking}
As discussed in Section \ref{coil design}, the pose of the coil above the scalp significantly influences both the distribution and intensity of the induced \gls{E-field} within the target stimulation region \cite{degoede2018accurate}. Tracking devices, a crucial component of the neuronavigation system \cite{glossop2009advantages}, enable accurate and real-time monitoring of the coil’s pose relative to the subject’s head. Conventional neurosurgical tracking methods include invasive frame-based systems, \bluetext{which secure pins to the subject’s skull before brain scanning and later attach rigid frames for navigation}, and non-invasive electromagnetic tracking, which localises sensors within a magnetic field. While both methods offer high accuracy, they are unsuitable for \gls{Robo-TMS} due to their operational inconvenience and susceptibility to interference. In contrast, optical-based frame-less stereotactic systems, also known as optical tracking systems, are widely employed in \gls{TMS}, relying on markers attached to the subject's head and the coil to track the coil’s pose relative to the head \cite{wagner2007noninvasive}.

Optical tracking systems can be classified into two types based on the light source: active and passive optical tracking. The key difference between them is whether the marker emits light. In active optical tracking, powered markers with integrated \glspl{LED} emit \gls{IR} light, which is tracked by \gls{IR} cameras to determine the pose of the instruments. This system is more resistant to ambient light interference, making detection easier. However, the markers require power, which introduces additional weight and complexity due to the need for batteries or wires. In contrast, passive optical tracking uses reflective markers attached to the subject or instruments. The optical camera detects light reflected from these markers to determine their pose. Since passive markers don’t require power, they enable simpler setups and are widely used.

Markers, also referred to as trackers, work alongside optical tracking cameras to form an optical tracking system. They are generally divided into three categories based on their form: array-based markers, pattern-based markers, and natural features (marker-less). Array-based markers, the most common in \gls{Robo-TMS}, consist of three or more reflective markers arranged in a known geometric configuration \cite{todd2014brain, noccaro2021development}. Their pose is captured by two or more cameras positioned at different angles, which triangulate the marker’s position. Pattern-based markers, on the other hand, use predefined, high-contrast patterns—such as AprilTags \cite{krogius2019flexible} and ArUco markers \cite{garrido-jurado2014automatic}—printed on flat surfaces \cite{souza2018development, lin2019trajectory}. Cameras detect and identify these unique patterns from a preloaded library and calculate the marker’s spatial pose. These markers are easier and cheaper to produce and manage and can operate under normal lighting conditions without relying on \gls{IR} light. However, both array-based and pattern-based systems require the attachment and management of external markers. Marker-less tracking, which relies on detecting natural features from the subject’s head and face, reduces setup time and eliminates the risk of marker detachment or movement during procedures \cite{z.xiao2018cortexbot, chen2020noattachment}. However, it is less robust compared to marker-based systems, primarily due to its relative immaturity and the challenges in consistently identifying natural surface features.

Conventionally, the optical camera is positioned outside the robot arm to maximise its \gls{FOV} for tracking the head and coil. However, this setup can result in tracking loss if the robot arm obstructs the camera’s line of sight during movement. To address this issue, an inside-out tracking method, where a portable camera is mounted on the robot’s end-effector, has been proposed \cite{liu2022out}. This method ensures that the camera maintains an unobstructed view of the head and coil throughout the robot’s motion.

\subsection{E-field Modelling}
Unlike neuronavigation systems for neurosurgery, which primarily track instrument poses relative to the brain, \gls{Robo-TMS} requires both real-time tracking of the coil pose and estimation of induced \gls{E-field} distribution to assess the electric currents that stimulate target neurons. The \gls{E-field} modelling is essential for understanding \gls{TMS} effects on neural tissue and tailoring stimulation protocols to specific brain regions or individuals, as \gls{TMS} needs to properly engage disease-related pathways to improve clinical outcome \cite{deng2020devicebased}. Accurate and real-time \gls{E-field} modelling relies on two factors: realistic head models and robust \gls{E-field} modelling solvers.

\subsubsection{Realistic Head Models}
Early head models in \gls{TMS} rely on simplified geometries as standard templates, such as infinite half-planes or perfect spheres fitted locally or globally to the subject’s head \cite{esselle1992neural, heller1992brain}. However, comparisons with more realistic head models \cite{wagner2004threedimensional, wagner2007noninvasive} have demonstrated that these simplified models lack accuracy due to the omission of structural details and tissue conductivity \cite{nummenmaa2013comparison, nieminen2022accuracy}. Modern head models, such as those developed by \cite{wei2019deep}, offer improved structural detail, resolution, and tissue conductivity, setting a new standard.

Despite improvements in standard head models, research highlights that individual anatomical variations significantly affect \gls{E-field} distribution, emphasising the need for personalised stimulation protocols \cite{laakso2015intersubject, laakso2018where}. In response, toolboxes have been developed to automate the creation of individualised head models from \gls{MRI} \cite{thielscher2015field, nielsen2018automatic, huang2019realistic}. These early toolboxes typically segment five major head tissues and assign different tissues with their conductivity for \gls{E-field} modelling \cite{thielscher2011impact, saturnino2018simnibs, huang2019realistic}. Further results in \cite{puonti2020accurate} demonstrate that increasing the number of segmented tissues to fifteen from \gls{MRI} significantly improves accuracy, reducing the relative error in \gls{E-field} estimation. This advancement allows for more precise and individualised \gls{E-field} modelling.

\subsubsection{E-field Modelling Solvers}
\gls{E-field} modelling solvers are computational tools used to estimate the \gls{E-field} induced by \gls{TMS} within the brain. These solvers estimate the \gls{E-field} distribution based on coil placement and current parameters, along with detailed head models that represent tissue geometry and conductivity. While high-resolution head models are crucial for accurate \gls{E-field} estimations in \gls{Robo-TMS}, they are often computationally expensive. To achieve real-time \gls{E-field} modelling, essential for \gls{TMS} treatment involving moving coils, the trade-off between computational speed and accuracy has to be considered \cite{gomez2020conditions}. Broadly, \gls{E-field} solvers can be categorised into two main types: physics-based solvers and learning-based solvers.

Basic physics-based solvers assume the \gls{E-field} peak occurs where the coil normally intersects the cortex \cite{herwig2001navigation}, enable real-time estimation but are inaccurate, particularly with non-tangential coil placements \cite{sollmann2016comparison}. More sophisticated physics-based solvers typically employ methods like \gls{FEM}, \gls{BEM}, and \gls{FDM}, which numerically solve Maxwell’s equations to estimate the \gls{E-field} induced by time-varying magnetic fields.

\gls{FEM} is widely used due to its ability to model complex geometries and varying material properties by dividing the head model into finite elements, such as tetrahedrons, and calculating electromagnetic fields locally. It is ideal for heterogeneous models but is computationally intensive due to the large number of elements required for high-resolution simulations \cite{bungert2016where, laakso2018where, wang2023fast, hasan2023realtime, hasan2023fast}. \gls{BEM}, on the other hand, simplifies computations by discretising only the model’s surfaces. It is particularly effective when the head can be approximated by layered tissues with homogeneous conductivity \cite{makarov2018quasistatic, htet2019comparative, stenroos2019realtime, daneshzand2021rapid, makaroff2023fast}. \gls{FDM} employs a structured grid to solve equations through finite difference approximations. While easier to implement due to structured grids, it is less flexible for complex geometries and better suited to simplified or idealised head models \cite{toschi2008reconstruction, laakso2012fast, paffi2015computational}. 

Physics-based solvers are widely adopted in \gls{E-field} modelling software. Commercial software—such as COMSOL Multiphysics \cite{silva2008elucidating}, ANSYS Maxwell 3D \cite{makarov2016preliminary, wei2019deep}, Sim4Life \cite{afuwape2021effect}, and SEMCAD X \cite{rastogi2017transcranial}—are not specifically developed for \gls{TMS} \gls{E-field} modelling, which limits their suitability for real-time \gls{Robo-TMS} applications. In contrast, open-source software like SimNIBS \cite{thielscher2015field, saturnino2018simnibs, puonti2020value, gomez2021fast} and ROAST \cite{huang2018roast, huang2019realistic} are well-maintained and facilitate visualisation of the \gls{E-field} distribution in the brain. Neural Navigator, a clinical software designed for \gls{nTMS} \cite{neggers2004stereotactic}, is compatible with various \gls{TMS} devices and has received \gls{FDA} clearance for clinical use.

In addition to these physics-based methods, learning-based \gls{E-field} estimation has emerged as a promising solution to reduce computation times in real-time modelling. These models leverage the computational power of \glspl{GPU} to learn and approximate complex relationships between coil placement, current parameters, head anatomy, and the resulting \gls{E-field} distribution during the training phase, enabling faster estimations \cite{yokota2019realtime, xu2021rapid, li2022computation, franke2024slicertms}. \bluetext{Current methods fall into two categories: supervised and self-supervised learning \cite{park2024review}. Supervised methods directly predict \glspl{E-field} by minimising error against reference data from physics-based \gls{FEM} simulations \cite{yokota2019realtime, xu2021rapid}, whereas self-supervised models predict electrical potential by optimising an energy function during training \cite{li2022computation}.} SlicerTMS, a recent extension of the open-source medical imaging platform 3D Slicer, delivers real-time \gls{E-field} modelling by leveraging learning-based solvers \cite{franke2024slicertms}.

\renewcommand\arraystretch{1.2}
\begin{table*}[!htb]
    \caption{The Comparison of Typical Benchmarks}
    \centering
    \resizebox{\linewidth}{!}{ 
        \begin{tblr}{
            hline{1,8} = {2pt},
            column{1} = {0.08\linewidth, valign=h, halign=c},
            column{2} = {0.26\linewidth, valign=h, halign=l},
            column{3} = {0.29\linewidth, valign=t, halign=l},
            column{4} = {0.29\linewidth, valign=t, halign=l},
            column{5} = {0.08\linewidth, valign=h, halign=c},
            row{1} = {valign=m, halign = c},
            stretch=-1,
            }
            \textbf{Benchmarks} & \textbf{Methods} & \textbf{Advantages} & \textbf{Disadvantages} & \textbf{References} \\ 
            \hline
            \textbf{Simulations} & 
            Simulations are computer-generated models that replicate neuronavigation processes under idealised conditions. They are often used in the early stages of system development to test and predict system performance, including spatial accuracy and precision. & 
            \begin{itemize}
                \item Fast and cost-effective for early testing.
                \item Can model a wide range of scenarios without requiring physical subjects or equipment.
                \item Provides a controlled environment, allowing developers to isolate and test specific variables.
            \end{itemize} & 
            \begin{itemize}
                \item May oversimplify real-world complexities, such as tissue deformation or sensor inaccuracies. 
                \item Does not account for human or biological variability. 
                \item Limited to theoretical validation, requiring further practical tests. 
            \end{itemize} & 
            \cite{deng2013electric, mikkonen2019effects, nieminen2022accuracy},\bluetext{\cite{caulfield2024mitigating}} \\
            \textbf{Phantoms} & 
            Phantoms are artificial models that replicate human anatomy, typically used to physically test neuronavigation systems. These models can range from simple geometric forms to highly detailed replicas of the skull, brain, or head tissues. & 
            \begin{itemize}
                \item Provides a realistic, hands-on way to validate system performance.
                \item Allows for repeated testing under consistent conditions.
                \item Avoids ethical concerns related to human trials.
            \end{itemize} & 
            \begin{itemize}
                \item May not fully replicate the complexities, such as heterogeneity and conductivity in live tissues. 
                \item Limited in representing the dynamic movements or changes that occur in real scenarios.
                \item Once constructed, phantoms are fixed and do not account for inter-subject variability.
            \end{itemize} & 
            \cite{richter2012handassisted, richter2013stimulus, noccaro2018evaluation, saturnino2019electric, jaroonsorn2020robotassisted, kebria2023hapticallyenabled, liu2025imageguided} \\
            \textbf{Landmark-based Coordinates} & 
            The landmark-based coordinate system, such as Talairach coordinates and the 10–20 system, is a standardised brain mapping framework based on anatomical landmarks.  & 
            \begin{itemize}
                \item Can handle partial individual differences, like brain size and overall shape.
                \item Well-established in clinical practice and easy to implement.
                \item Non-invasive and safe.
            \end{itemize} & 
            \begin{itemize}
                \item Template-based and may not accurately reflect individual anatomical differences. 
                \item Highly dependent on operation, introducing subjective errors.
                \item Low resolution.
            \end{itemize} & 
            \cite{young2022comparison} \\
            \textbf{Optical Tracking} & 
            Optical tracking systems use cameras to track the pose of markers attached to the subject's head or tools in real-time. It can be employed to compare errors of various localisation methods for the same hotspot.  & 
            \begin{itemize}
                \item Provides real-time pose feedback during navigation.
                \item Widely used in clinical practice, making it an established method.
                \item Non-invasive, and compatible with a range of clinical setups.
            \end{itemize} & 
            \begin{itemize}
                \item Requires a direct line of sight between the camera and markers during procedures.
                \item Markers may shift during procedures, introducing errors.
                \item Difficult to locate the error of the system itself due to external factors like lighting or reflections.
            \end{itemize} & 
            \cite{carducci2012accuracy, degoede2018accurate, goetz2019accuracy, chen2020noattachment, caulfield2022neuronavigation, dormegny-jeanjean20223dmapping, shin2023robotic, xygonakis2024transcranial} \\
            \textbf{3D Scans} & 
            3D scans from \gls{MRI}, \gls{CT}, depth cameras and lidars, represent the head and tool within a unified observation space, enabling direct comparison of actual relative poses with ideal reference values. & 
            \begin{itemize}
                \item Provides highly detailed anatomical information and accurate relative poses.
                \item Non-invasive.
            \end{itemize} & 
            \begin{itemize}
                \item Cannot be processed in real-time.
                \item May involve complex post-processing steps, like aligning the scan with the anatomy.
                \item Device-dependent and can be expensive.
            \end{itemize} & 
            \cite{hironaga2019proposal, matilainen2024verification} \\
            \textbf{Direct Electrical Cortical Stimulation (DECS)} & \gls{DECS} is a technique in which electrical impulses are applied directly to the brain during surgery to study the relationship between cortical structure and systemic function, which provides real-time feedback by observing motor or sensory responses to stimulation. & 
            \begin{itemize}
                \item Considered the gold standard for accuracy as it provides physiological confirmation of navigation.
                \item Provides direct, real-time validation in live subjects.
            \end{itemize} & 
            \begin{itemize}
                \item Invasive, requiring an open-skull procedure and limited to intraoperative settings.
                \item Only applicable in highly specific clinical cases, primarily during neurosurgery.
                \item Ethical and practical limitations prevent it from being a routine validation tool.
            \end{itemize} & 
            \cite{tarapore2012preoperative, krieg2012utility, takahashi2013navigated, jeltema2021comparing, indharty2023comparison, muscas2023headsmicronavigation} \\
        \end{tblr}
    }
    \label{tab:comparison of benchmarks}
\end{table*}

\subsection{Benchmarks and Metrics}
The primary performance metrics for neuronavigation systems in \gls{Robo-TMS} are accuracy and precision. Accuracy measures how closely the actual stimulation spot aligns with the intended brain target, typically quantified as the Euclidean distance and relative angle between the target and the actual stimulation spot, expressed in \(mm\) and \(^\circ\). Precision refers to the system’s ability to consistently return to the same target across multiple attempts, also measured in \(mm\) and \(^\circ\), usually as the standard deviation of repeated localisations at the same location. Precision is essential for \gls{rTMS}, where repeated targeting is required over multiple sessions. While accuracy is about hitting the target spot, precision is about consistently hitting the same spot. A system can be precise but not accurate if it consistently deviates from the target by a fixed offset. Accuracy is often influenced by registration method, imaging quality in optical tracking, and the use of individualised \glspl{MRI}, while precision depends on calibration method, spatial consistency of the optical tracking system, and repeatability of the robot’s movements \cite{nieminen2022accuracy}.

Evaluating spatial accuracy and precision in neuronavigation systems remains challenging, particularly in living tissues where uncertainties persist. Research often focuses on optimising specific aspects of accuracy and precision under controlled conditions and uses different benchmarks (simulation, phantoms, etc) for comparison. The TABLE \ref{tab:comparison of benchmarks} lists typical benchmarks in current literature, facilitating comparative analysis of spatial accuracy and precision among methods.

Additional performance metrics relevant to neuronavigation systems in \gls{Robo-TMS} include calibration and registration time \cite{shin2023robotic}, real-time performance of \gls{E-field} modelling, effective tracking range like \gls{FOV}, and user satisfaction gathered through surveys. These supplementary criteria offer a broader perspective on system performance in clinical contexts, providing a more comprehensive foundation for evaluating advancements in neuronavigation systems.

\section{Control Systems}
The goal of the \gls{Robo-TMS} control system is to meet two primary clinical requirements: compensating for head movement to maintain the stimulation target in real-time, and sustaining a consistent coil-to-head contact force to ensure reliable contact while minimising the risk of discomfort or injuries.

\subsection{Motion Compensation}
In conventional \gls{TMS} setups, the coil is mounted on a static holder, and subjects are instructed to avoid head movement, often with the aid of head restraints \cite{chronicle2005development}. In contrast, \gls{Robo-TMS} allows head movement during treatment sessions hence improving comfort. To maintain the spatial relationship between the coil and the head and achieve accurate stimulation, the robot needs to actively compensate for the subject’s head movement. This spatial relationship between the \gls{TMS} coil and the subject’s head can be represented using a homogeneous transformation matrix:
\begin{equation}
\mathbf{T} = \begin{bmatrix}
\mathbf{R} & \mathbf{t} \\
\mathbf{0}^\top & 1
\end{bmatrix}
\in \mathbb{R}^{4\times4}
\end{equation}
where \(\mathbf{T}\) lies in the special Euclidean group \(SE(3)\), \(\mathbf{R}\) is a rotation matrix in the special orthogonal group \(SO(3)\) and \(\mathbf{t} \in \mathbb{R}^3\) is a translation vector. We denote transformations from one frame to another as  \(^A\mathbf{T}_B\), representing a transformation that converts coordinates in frame \(\{B\}\) to frame \(\{A\}\). Accordingly, the relationship between the \gls{TMS} coil and the subject’s head can be represented as \(^{head}\mathbf{T}_{coil}={^{coil}\mathbf{T}_{head}}^{-1}\). 

The goal of \gls{Robo-TMS} can then be expressed in terms of the clinical requirement for stimulation targeting as:
\begin{equation} \label{equation: head to coil requirement}
^{head}\mathbf{T}^*_{coil} = ^{head}\mathbf{T}_{MRI} \cdot ^{MRI}\mathbf{T}^*_{coil}
\end{equation}
where \(\{head\}\) is the subject's head frame, \(\{coil\}\) is the \gls{TMS} coil frame, and \(\{MRI\}\) is the 3D \gls{MRI} frame. The transformation \(^{head}\mathbf{T}_{MRI}\) is obtained through registration, while \(^{MRI}\mathbf{T}^*_{coil}\) specifies the clinical requirement by defining the stimulation target clinicians aim to place the coil on.

In contrast to this clinical requirement in \eqref{equation: head to coil requirement}, the control objective is defined by an alternative expression:
\begin{equation} \label{equation: head to coil control}
^{head}\mathbf{T}^*_{coil} = ^{head}\mathbf{T}_{cam} \cdot ^{cam}\mathbf{T}_{robot} \cdot ^{robot}\mathbf{T}^*_{end} \cdot ^{end}\mathbf{T}_{coil}
\end{equation}
where \(\{cam\}\) is the optical tracking camera's base frame, \(\{robot\}\) is the robot's base frame, \(\{end\}\) is the robot's end-effector frame. Here, \(^{cam}\mathbf{T}_{robot}\) and \(^{end}\mathbf{T}_{coil}\) are determined during calibration, while \(^{head}\mathbf{T}_{cam} = {^{cam}\mathbf{T}_{head}}^{-1}\) is acquired in real-time from the optical tracking system. As a result, \(^{robot}\mathbf{T}^*_{end}\) represents the real-time control target, which can be calculated by solving equations \eqref{equation: head to coil requirement} and \eqref{equation: head to coil control} in the given \(^{head}\mathbf{T}^*_{coil}\) \cite{lin2019trajectory, noccaro2021development}.

For cases where the optical tracking system cannot be fixed in the operating room, motion compensation can be reformulated as a tracking problem. At the cost of maintaining a line of sight to both the head and end-effector markers throughout the procedure, this method requires only the transformation \(^{coil}T_{end}\) to be calibrated in advance \cite{xygonakis2024transcranial}. The formulation is given by:
\begin{align}
    & ^{cam}\mathbf{T}_{coil} \cdot ^{coil}\mathbf{T}_{end} \cdot ^{end}\mathbf{T}_{robot} \nonumber \\
  = & ^{cam}\mathbf{T}_{head} \cdot ^{head}\mathbf{T}^*_{coil} \cdot ^{coil}\mathbf{T}_{end} \cdot ^{end}\mathbf{T}^*_{robot}
\end{align}

where \(^{coil}\mathbf{T}_{end}\) is pre-determined via calibration, \(^{cam}\mathbf{T}_{head}\) and \(^{cam}\mathbf{T}_{coil}\) are obtained in real-time from optical tracking, and \(^{end}\mathbf{T}_{robot}\) is provided by the robot arm. The control target \(^{end}\mathbf{T}^*_{robot}\) is then solved using the clinical requirement \(^{head}\mathbf{T}^*_{coil}\). This method eliminates the need for \(^{cam}\mathbf{T}_{robot}\), allowing the use of a mobile camera.

\subsection{Contact Force Control}
Sustaining a consistent coil-to-head contact force is another essential clinical requirement for \gls{Robo-TMS}, as it ensures reliable contact while minimising the excessive pressure which may cause discomfort or injuries to the subject. As the absence of force feedback is a known limitation \cite{goetz2019accuracy}, contact force control is strongly recommended in \gls{Robo-TMS}. Integrating a force or torque sensor between the robot's end-effector and the coil allows \gls{Robo-TMS} to monitor and regulate the coil’s contact with the head surface, ensuring sufficient contact without discomfort or injuries. To further improve force accuracy, one study estimates the shape of the deformable coil cable to compensate for the dynamic forces applied to the robot’s end-effector \cite{richter2013robotized, zhang2024realtime}.

Accurately modelling the subject’s head surface is challenging due to variations in skull shape, hair volume, and underlying tissue composition. Additionally, the subject’s head remains unconstrained during the \gls{TMS} procedure. Sustaining a consistent coil-to-head contact force is essential to ensure reliable contact under modelling and motion uncertainties while minimising the excessive pressure which may cause discomfort and injuries. Two main force control methods adopted in \gls{Robo-TMS} are impedance control \cite{g.pennimpede2013hot, noccaro2021development} and force/position hybrid control \cite{lin2018dynamic, jaroonsorn2020robotassisted}. For both control methods, current research typically adopts a two-stage control strategy: an initial free-space movement phase when the coil is relatively far from the subject's head, followed by an intraoperative phase that maintains a force normal to the subject’s head surface (constrained-space movement). A study indicates that a normal force of less than \(10 \mathrm{N}\) is typically applied to keep the coil in close contact with the head surface without causing discomfort \cite{zakaria2012forcecontrolled}. \bluetext{In future developments, exploring advanced force control strategies used in other surgical robotic systems, such as model predictive control with force feedback \cite{dominici2014model} and learning-based force control \cite{ren2021learning}, may enhance adaptability and robustness under anatomical variability and motion uncertainty.}

\subsection{Safety Assurance}
In industrial applications, safety can be guaranteed by keeping operators outside the robot’s workspace or by halting the system if a person approaches too closely. However, in \gls{Robo-TMS}, subjects must remain within the robot’s workspace. The need for close interaction with the human environment makes \gls{Robo-TMS} design uniquely challenging, as any system failure could be critical. For safe robot-human interaction, the system must operate with high reliability and adhere to strict safety constraints.

Safety requirements for \gls{Robo-TMS} can be divided into two levels: hardware and software. On the hardware level, researchers typically incorporate safety norms by adding power buttons, emergency stops, start buttons, and collision sensors. Design considerations also include limiting the robot’s movement to a confined workspace through specialised structural integration \cite{zorn2012design}. On the software level, because trajectory planning and execution remain complex, errors may occur during operation to cause harmful collisions for subjects. To mitigate risks, motion limits and monitoring systems are essential, including speed limits, joint limits, contact force monitoring, movement tracking, and watchdog timers \cite{kantelhardt2010robotassisted, rossi2021safety}.

\section{Discussion}\label{discussion}
\gls{Robo-TMS} enhances conventional \gls{TMS} by integrating advanced robotics, enabling more accurate and repeatable stimulation targeting, which streamlines the treatment process and significantly increases its efficacy. Despite significant progress, \gls{Robo-TMS} remains in its early stages, with a noticeable gap between the development of robotic technologies and clinical requirements. Furthermore, these advancements introduce new concerns that require careful consideration. This section explores the clinical implications, current challenges, and future directions, focusing on how to bridge the gap between medical demands and engineering innovations.

\subsection{Clinical Implications}
In this part, we will discuss the application of \gls{Robo-TMS} in a clinical context, highlighting the importance of keeping clinical scenarios in mind when integrating cutting-edge robotic technologies.

\subsubsection{Reproducible Treatment}
The \gls{Robo-TMS} system significantly enhances clinical performance by delivering highly accurate and repeatable stimulation to specific brain regions, which is crucial for targeting small or deep brain structures\bluetext{\cite{ginhoux2013custom, dormegny-jeanjean20223dmapping}}. Unlike conventional \gls{TMS}, which suffers from operator variability, robotic systems provide reproducible stimulation across sessions, improving clinical outcomes. Integrated optical tracking systems compensate for head movements, ensuring accurate stimulation throughout \gls{rTMS} procedures \cite{shin2023robotic}. Additionally, neuronavigation systems enable customised protocols tailored to each subject's neuroanatomy, maintaining consistent efficacy over long-term treatment courses and further enhancing therapeutic reliability and quality of care\bluetext{\cite{richter2013optimal, degoede2018accurate}}.

\subsubsection{Accurate Brain Mapping}
The high targeting accuracy of \gls{Robo-TMS}, typically within a few millimetres, incorporates factors such as robot arm flexibility and neuronavigation accuracy. This capability has made \gls{Robo-TMS} a valuable tool for brain mapping, facilitating the study of brain structure-function relationships \cite{grab2018robotic, biernacki2020recovery, giuffre2021reliability, kahl2022active, kahl2023reliability}. Clinically, brain mapping is increasingly employed for preoperative motor and language function mapping in subjects with brain tumours \cite{lefaucheur2016value, jeltema2021comparing}. Although language mapping tends to show lower concordance with intraoperative cortical mapping compared to motor mapping \cite{haddad2020preoperative}, \gls{TMS}-guided mapping has been shown to reduce postoperative deficits with improved surgical outcomes, particularly when tumours near motor pathways\bluetext{\cite{frey2014navigated, picht2016presurgical, raffa2019role}}.

\subsubsection{Reduced Side Effects}
The accuracy of \gls{Robo-TMS} significantly reduces the risk of side effects by targeting specific regions without affecting unintended regions. This contrasts with non-focal electroconvulsive therapy, where non-specific stimulation frequently leads to generalised seizures \cite{rossi2021safety}. In a study of 733 brain tumour subjects, including 50\% with a seizure history, motor and language mapping with neuronavigation systems was found to be safe and well-tolerated \cite{tarapore2016safety}. Specifically, researchers used the \gls{VAS} to assess discomfort (VAS 1-3) and pain (VAS \(>\)3) and found that discomfort occurred in only 5.1\% of motor mapping cases and 23.4\% of language mapping cases, with minimal pain during motor mapping (0.4\%) but higher pain rates during language mapping (69.5\%). Despite the high seizure risk of the subjects, no seizures were observed during the procedures, and no \gls{nTMS}-related seizures had been reported elsewhere \cite{rossi2021safety}, highlighting its safety for motor and language mapping.

\subsubsection{Improved Clinician Support}
In conventional \gls{TMS}, clinicians manually place the coil and continuously monitor its pose throughout the session, leading to increased fatigue and stress\bluetext{\cite{xygonakis2024transcranial}}. Repeated exposure to magnetic fields and frequent physical interaction between clinicians and subjects raise concerns about long-term safety and infection transmission risks \cite{rossi2021safety}. By automating coil placement and stimulation delivery, \gls{Robo-TMS} can alleviate clinician fatigue and stress. By reducing direct contact, it minimises magnetic field exposure and lowers the risk of human-to-human infection transmission\bluetext{\cite{bikson2020guidelines}}. The integration of robotic technologies not only optimises clinical outcomes but also improves working conditions for clinicians.

\subsubsection{Potential Discomfort for Subjects}
While it offers significant benefits, \gls{Robo-TMS} may also introduce potential discomfort for subjects \cite{ginhoux2013custom, peterchev2015advances, tarapore2016safety}. A common concern in all \gls{rTMS} treatments is the noise from the coil, which can be particularly unsettling for some individuals\bluetext{\cite{edwards2024practical}}. For instance, sound pressure levels can exceed \(76 \mathrm{dB(A)}\) at \(25 \mathrm{cm}\) during a \(20 \mathrm{Hz}\) pulse train at maximum stimulator output \cite{koponen2020sound}. Additionally, the presence of robots may cause fear or anxiety in individuals unfamiliar with robotic systems. Moreover, the continuous pressure exerted by the coil on the subject’s head can lead to musculoskeletal discomfort \cite{rossi2021safety}. The headband, which secures optical tracking markers, may further cause irritation or discomfort during prolonged sessions\bluetext{\cite{omara2014anatomical}}. Although these factors are secondary to the primary clinical objectives, addressing them is crucial for enhancing subject comfort and preventing premature termination of \gls{Robo-TMS} treatments.

\subsection{Current Challenges}
Despite advancements in \gls{Robo-TMS}, its clinical adoption remains limited, highlighting a gap between medical needs and engineering innovations. Accessibility is identified as the foremost barrier to broader clinical adoption, driven by unverified clinical applicability, high operational complexity, and substantial implementation costs.

\subsubsection{Unverified Clinical Applicability}
Although \gls{Robo-TMS} aims to deliver more accurate and repeatable stimulation, the clinical significance of this advantage remains unclear\bluetext{\cite{romero2019neural, fitzgerald2021targeting}}. The complex and poorly understood relationships among \gls{TMS} stimulation, brain structure, neurological function, and clinical outcomes make it difficult to define how accurate the system needs to be\bluetext{\cite{passera2022exploring, siebner2022transcranial}}. The lack of fine-scale brain atlases and comprehensive neuroscientific models further complicates target selection for effective therapy \cite{deng2020devicebased}. Moreover, it remains to be verified in clinical practice whether \gls{Robo-TMS} offers substantial clinical benefits over conventional \gls{TMS} procedures. Evidence-based guidelines are needed to clarify when and how \gls{Robo-TMS} should be used. Finally, the engineering assumption of head rigidity limits applicability \cite{risholm2011multimodal, orringer2012neuronavigation}, excluding subjects with involuntary facial movements and narrowing the scope of potential subjects.

\subsubsection{High Operational Complexity}
Excessive manual intervention in calibration and registration is a major contributor to the high operational complexity of \gls{Robo-TMS}\bluetext{\cite{xygonakis2024transcranial}}. Although these steps are critical for maintaining stimulation efficacy, they are among the least automated aspects of the system. Workspace calibration must be repeated whenever the stimulation coil is changed or the operating room is rearranged. Manual registration involves identifying corresponding landmarks on the subject’s head and \gls{MRI}, making it time-consuming and heavily reliant on operator expertise\bluetext{\cite{omara2014anatomical, gao2024individualized}}. Stimulation intensity calibration must also be performed individually, as it depends on each subject’s neuroanatomy. These labour-intensive procedures are prone to subjective errors, which may compromise efficacy. Moreover, reliance on markers introduces further complexity, as any displacement during treatment often necessitates re-calibration or re-registration, thereby extending the overall workflow \cite{smith201630}.

\subsubsection{Substantial Implementation Costs}
High implementation costs remain a major barrier to the widespread clinical adoption of \gls{Robo-TMS}. Advanced robotic platforms, optical tracking systems, and precise neuronavigation tools require substantial upfront investment and ongoing maintenance\bluetext{\cite{matsuda2024robotic, liu2025imageguided}}. Additionally, the cost of acquiring \glspl{MRI} for registration and training clinicians and technicians further inflates expenses. Although cost-effective alternatives, such as personalised helmet-based coil localisation \cite{wang2022individualized}, have been proposed in clinical practice, they often compromise flexibility and adaptability. These challenges highlight the pressing need to reduce system complexity and operational costs to support broader deployment.

\begin{figure}[!htb]
\centering
\includegraphics[width=0.48\textwidth]{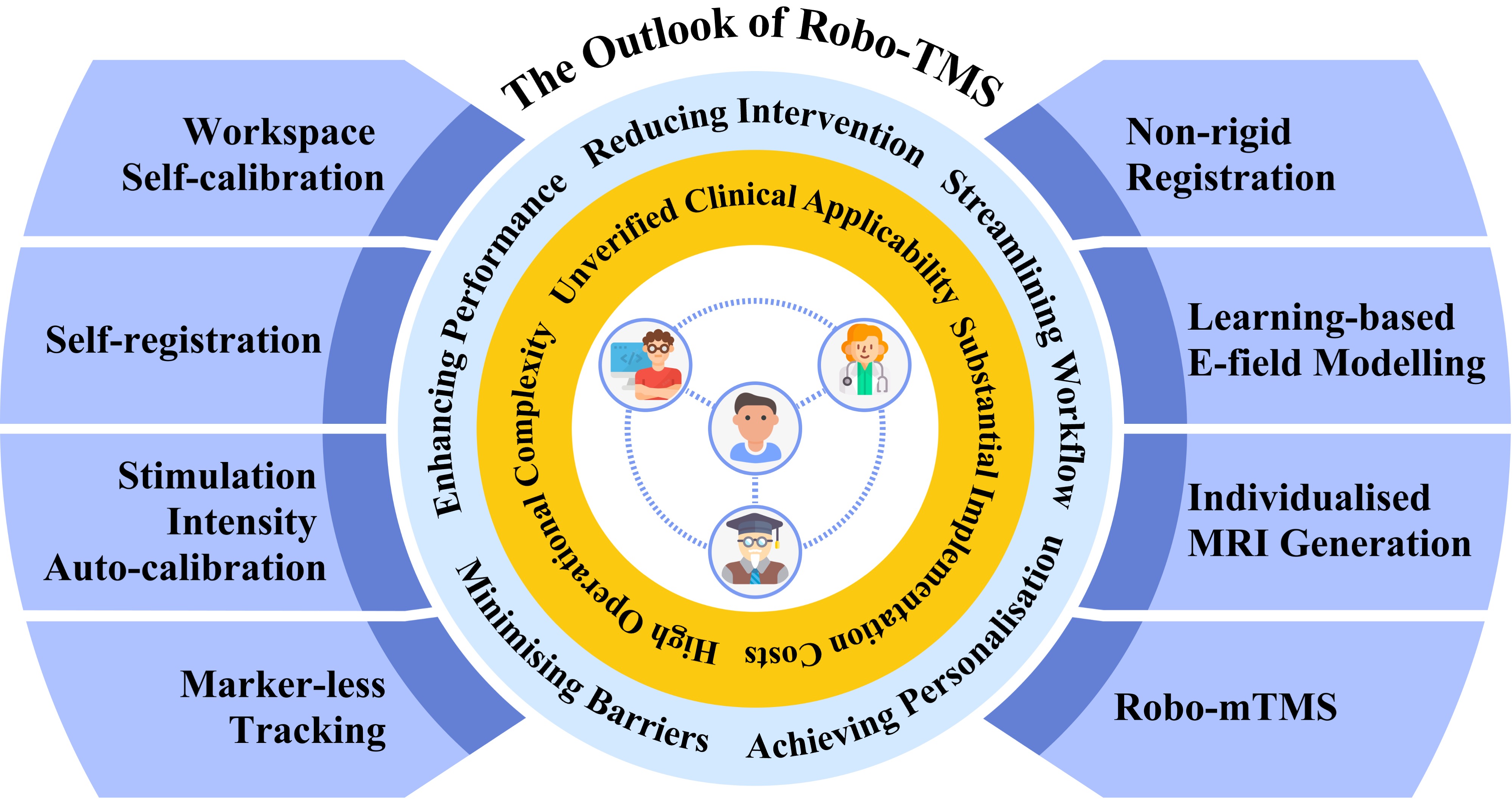}
\caption{The outlook of \gls{Robo-TMS}. Advancements in \gls{Robo-TMS} have been driven by collaborative efforts among clinicians, engineers, and researchers. However, the challenges highlighted in the yellow ring remain unresolved, hindering progress towards the future directions outlined in the light blue ring. Promising solutions have begun to emerge, requiring further exploration to realise their full potential.}
\label{fig:outlook}
\end{figure}

\subsection{Future Directions}
The key technical objectives of \gls{Robo-TMS} are achieving greater accuracy, repeatability, and efficiency, while improving user-friendliness and lowering costs are prerequisites for large-scale adoption. To address these priorities, the following future directions can be explored.

\subsubsection{Reducing Manual Intervention}
While many steps in \gls{Robo-TMS} have been automated, significant manual intervention remains in calibration and registration. These steps are time-consuming and heavily reliant on the operator's expertise, limiting both efficiency and accuracy\bluetext{\cite{nieminen2022accuracy, koehler2024how}} \bluetext{while inflating training costs \cite{smith201630}.} Advancements in robotics offer promising solutions to minimise manual intervention. For instance, workspace self-calibration could eliminate the assumption of fixed transformations between the camera and robot base frames, enabling the system to adapt to alignment shifts or allowing clinicians to move the camera for optimal views during procedures \cite{xygonakis2024transcranial}. Similarly, self-registration methods could utilise natural facial features for automated head-to-MRI alignment, thereby reducing manual errors\bluetext{\cite{liu2023transcranial}}. Incorporating non-rigid registration methods could further relax the need for subjects to replicate their expressions during \gls{MRI} scanning, minimising the effort required to ensure proper alignment\bluetext{\cite{risholm2011multimodal, orringer2012neuronavigation}}.

\subsubsection{Enhancing System Performance}
Enhancing accuracy, repeatability, and real-time performance remains a core goal in \gls{Robo-TMS}. These improvements are intrinsically linked to the capabilities of neuronavigation systems. Insights from research in both \gls{Robo-TMS} and advanced robotics highlight promising developments. For instance, surface-based registration accuracy can be improved by collecting more scalp data points, paving the way for more automated solutions \cite{noirhomme2004registration, nieminen2022accuracy}. Marker-less tracking methods, employing 3D laser scanning, TOF-based cameras, or structured-light cameras, directly track head movements without relying on conventional stereo cameras and head markers\bluetext{\cite{richter2011navigated, chen2020noattachment, liu2023transcranial}}. This method not only eliminates errors associated with markers and digitiser tools but also enhances subject comfort by removing the need for headbands, offering improved accuracy in specific scenarios \cite{hironaga2019proposal}. Furthermore, learning-based \gls{E-field} modelling significantly enhances real-time performance, further advancing the ability to deliver immediate and effective treatments\bluetext{\cite{park2024review}}.

\subsubsection{Achieving Personalised Treatment}
Stimulation intensities in \gls{TMS} are typically expressed as a percentage of \gls{MT} to ensure safety and standardisation across subjects and stimulation spots \cite{herbsman2009motor}. However, this percentage-based method can be vague and indirect, as it fails to fully account for individual neuroanatomical variations, differences in stimulation spots, and the diversity of coils and devices used\bluetext{\cite{caulfield2024mitigating}}. Future advancements in neuroanatomy exploration and accurate \gls{E-field} modelling  \cite{dannhauer2024electric} may allow for stimulation intensities to be defined in absolute units, such as current intensity \(\mathrm{mA}\), enabling more accurate subject-specific dosages. Moreover, automating the stimulation intensity calibration in \gls{Robo-TMS} could further minimise subjective variability and achieve repeatable treatment\bluetext{\cite{harquel2017automatized}}.

\subsubsection{Minimising Implementation Barriers}
Clinicians often rely on individual \glspl{MRI} for accurate registration in \gls{Robo-TMS}, but their high cost and the need for specialised equipment pose barriers\bluetext{\cite{edwards2024practical}}. Additionally, it is uncommon for subjects to undergo expensive \gls{MRI} scans in conventional \gls{TMS} treatments. Emerging brain mapping methods offer a promising alternative by generating individualised \glspl{MRI} from group or average data with sufficient accuracy \cite{carducci2012accuracy, gao2024individualized}. This innovation eliminates the need for \gls{MRI} scans prior to procedures, thereby minimising implementation barriers. Moreover, while industrial robot-based \gls{Robo-TMS} systems have become more accessible due to the widespread adoption of industrial robots\bluetext{\cite{richter2013robotized, noccaro2021development, xygonakis2024transcranial}}, the expansion of the \gls{Robo-TMS} market is expected to drive the re-emergence of specialised systems with enhanced clinical features\bluetext{\cite{zorn2012design, ginhoux2013custom}}. \bluetext{The broader adoption of either industrial or specialised systems is also expected to further reduce costs through economies of scale.}

\subsubsection{Streamlining Stimulation Workflow}
In \gls{Robo-TMS}, even minor movement or change in stimulation spots require physically moving the coil, which can be slow due to the robot arm's mechanical limitations and the need for accurate localisation near the scalp to ensure efficacy and safety. Advanced integration, such as robot-assisted \gls{mTMS}, overcomes these challenges by using \gls{mTMS} to adjust stimulation spots rapidly within small regions without moving the coil, while the robot arm can make slower adjustments when needed for larger regions \cite{matsuda2024robotic}. Combining the capabilities of \gls{Robo-TMS} with \gls{mTMS}, such integration potentially improves repeatability, reduces treatment duration, and enhances subject comfort by streamlining the stimulation workflow \cite{sinisalo2024modulating}.

\section{Conclusion}\label{conclusion}
\gls{Robo-TMS} shows great promise in enhancing the accuracy and repeatability of conventional \gls{TMS} treatment; however, its clinical adoption remains limited, highlighting a gap between medical needs and engineering innovations. This review systematically analyses four critical aspects—hardware and integration, calibration and registration, neuronavigation systems, and control systems—to identify current engineering challenges linking medical requirements. Accessibility is identified as the foremost barrier of \gls{Robo-TMS}, driven by unverified clinical applicability, high operational complexity, and substantial implementation costs. Emerging technologies, including marker-less tracking, non-rigid registration, learning-based \gls{E-field} modelling, individualised \gls{MRI} generation, \gls{Robo-mTMS}, and automated calibration and registration, present promising pathways to address these challenges, hence could potentially facilitate the clinical translation and broader adoption of \gls{Robo-TMS}.

\FloatBarrier


\bibliographystyle{IEEEtran}
\ifthenelse{\boolean{showcolor}}%
    {

}%
    {\bibliography{references}}%


\end{document}